\documentclass[letterpaper, 10 pt, conference]{ieeeconf}  

\IEEEoverridecommandlockouts                              

\usepackage{amsmath} 
\usepackage{amssymb}  
\usepackage{graphicx}
\usepackage{subcaption}
\usepackage[colorlinks]{hyperref}
\usepackage{graphics}
\usepackage{xspace}
\usepackage{multirow}
\usepackage{tabularx}
\usepackage{stfloats}
\usepackage{cite}

\newcommand{\ie}{\textit{i.e.}\xspace}
\newcommand{\eg}{\textit{e.g.}\xspace}

\usepackage{etoolbox}
\AtBeginEnvironment{tabular}{\footnotesize}

\usepackage{booktabs,siunitx,caption} 
\usepackage{color}
\usepackage{xcolor}

\definecolor{darkergreen}{RGB}{0,200,0}

\title{\LARGE \bf
Probabilistic Uncertainty Quantification of Prediction Models with Application to Visual Localization
}

\author{Junan Chen$^{*\dagger}$, Josephine Monica$^{*\dagger}$, Wei-Lun Chao$^{\ddag}$, and Mark Campbell$^{\dagger}$
\thanks{$^*$Equal contributions}
\thanks{$^{\dagger}$Mechanical and Aerospace Engineering Department, Cornell University 
        \tt\small {\{jc3342, jm2684, mc288\} @cornell.edu}}%
\thanks{$^{\ddag}$Department of Computer Science and Engineering, the Ohio State University \tt\small{chao.209@osu.edu}}%
}
\newcommand{\cone}[1]{\textcolor{violet}{#1}} 
\newcommand{\ctwo}[1]{\textcolor{blue}{#1}} 
\newcommand{\cthree}[1]{\textcolor{purple}{#1}}

\begin{document}

\maketitle

\begin{abstract}
The uncertainty quantification of prediction models (\eg, neural networks) is crucial for their adoption in many robotics applications.
This is arguably as important as making accurate predictions, especially for safety-critical applications such as self-driving cars. 
This paper proposes our approach to uncertainty quantification in the context of visual localization for autonomous driving, where we predict locations from images.
Our proposed framework estimates \emph{probabilistic} uncertainty by creating a \emph{sensor error model} that maps an internal output of the prediction model to the uncertainty. 
The sensor error model is created using \emph{multiple} image databases of visual localization, each with ground-truth location.
Moreover, we integrate Gaussian Mixture Models (GMMs) to enhance the sensor error model, resulting in a more precise representation of uncertainty.
We demonstrate the accuracy of our uncertainty prediction framework using the Ithaca365 dataset, which includes variations in lighting, weather (sunny, snowy, night), and alignment errors between databases.
We analyze both the predicted uncertainty and its incorporation into a Kalman-based localization filter. Our results show that prediction error variations increase with poor weather and lighting condition, leading to greater uncertainty and outliers, which can be predicted by our proposed uncertainty model. 
\end{abstract}
\section{Introduction}
\label{sec:introduction}


The evolution of modern prediction models (\eg, neural networks) has revolutionized the performance of applications ranging from medical diagnostics, business analysis to robotics. 
However, much of the research in this field has focused primarily on enhancing performance (\eg, average prediction accuracy) through better data collection and architectures.
Despite these advancements, one significant weakness of many models is their inability to provide a sense of confidence in individual predictions. 
Predictive accuracies of these models can vary based on factors such as the amount and diversity of training data, the model architecture details, and the complexity of the test environment~\cite{wang2020train,CyCADA2017}.

In certain applications, such as medical imaging or self-driving, \emph{probabilistic uncertainty quantification} of prediction outputs is crucial. Realizing uncertainty models for these networks will not only facilitate their integration into formal probabilistic perception and planning frameworks but also enable better reasoning over the outputs. For example, in medical diagnosis, doctors should intervene when the neural network lacks confidence in its prediction~\cite{Jiang}. While some modern neural networks attempt to output probabilistic uncertainty, the reliability of the uncertainty prediction is still insufficient for safety-critical decision-making~\cite{NEURIPS2021_61f3a6db}. Most modern neural networks are deterministic or produce only \emph{non-probabilistic} confidence, such as the softmax function.


Current uncertainty modeling methods can generally be divided into three categories: Bayesian neural networks, ensemble, and post-processing methods. 
Bayesian neural networks \cite{10.1162/neco.1992.4.3.448,10.5555/525544} construct an inherent uncertainty estimation framework by formalizing a probability distribution over the model parameters~\cite{Gawlikowski2021ASO}. 
However, they are difficult to train and often output poorly calibrated confidence scores \cite{pmlr-v80-kuleshov18a}.
Ensemble methods~\cite{10.5555/3295222.3295387} typically train multiple neural networks with different training data or architectures, and the variance of the networks' output can indicate the uncertainty level. 
However, these methods require larger networks and additional training and inference steps.
Post-processing methods, such as neural network calibration, are general enough to be used with different networks. 
However, they require uncalibrated uncertainty as an input and cannot predict uncertainty directly. 
Examples include  histogram binning~\cite{Zadrozny2001ObtainingCP} and isotonic regression~\cite{Zadrozny}. 
Some post-processing methods, such as Platt scaling~\cite{Platt99probabilisticoutputs}, can predict uncertainty directly but require additional layers to be trained. The output of these methods is typically a simple confidence score, which is calibrated to be an approximate probability of correctness.

This paper presents a \emph{general} uncertainty prediction framework that does not require additional training of the network or changes in network architecture. 
The framework is \emph{probabilistically} formulated to provide both probability/confidence and an uncertainty distribution across the outputs. 
To achieve this, we leverage the concept of \emph{sensor models} in estimation frameworks (\eg, Kalman filter). 
For traditional sensors, manufacturers typically provide error model specifications that indicate the accuracy of the sensor under different conditions, \eg the accuracy of LiDAR as a function of range or the covariance of pseudo-ranges for GPS in various weather conditions. 
We propose creating an error uncertainty model for the network predictions using the internal network outputs and analysis across datasets. 

We demonstrate the effectiveness of our uncertainty prediction approach using the problem of visual localization~\cite{8269404}. 
We focus on this problem for two reasons: first, the neural network outputs a 2D position from an image, making it easy to analyze, and second, the network's performance is known to degrade in poor weather and lighting conditions\cite{Kapture2020,DS3M2020}. 
We build upon a typical visual localization model~\cite{8953492} which predicts the pose of a query image by searching the most similar image from a database of images with known poses using keypoint matching~\cite{superglue}.
Firstly, we analyze the performance of a baseline neural network to understand its performance over different databases (weather and lighting). 
We then create a statistical error model using the internal outputs of the network (number of keypoint matches between the query and retrieved images) as the \emph{cue} to predict visual localization error uncertainty. Importantly, the matched keypoints of each model/database can be calibrated and binned based on both a probability and 2D error. During inference, given the number of keypoint matches from an image, the sensor error model can directly return an uncertainty estimate in the form of a 2D error covariance (analogous to a traditional sensor) and a formal confidence. Moreover, we delve into the integration of Gaussian Mixture Models (GMMs) into the sensor error model, employing them to encapsulate uncertainty through Gaussian mixtures.
We can also incorporate the error model output in a Kalman-based localization filter, which provides a range of formal evaluation tools such as filter integrity and sensor hypothesis testing. 
 We evaluate our approach using Ithaca365~\cite{Diaz-Ruiz_2022_CVPR}, a large-scale real-world self-driving dataset that includes \emph{multiple} traversals along repeated routes, varying weather and lighting conditions, and high precision GPS. 

%

Our main contributions are four-folded: First, we analyze a state-of-the-art neural network for visual localization across a comprehensive dataset that includes multiple routes, lighting, and weather conditions to understand how errors vary across these key conditions. Second, we propose an approach to predict well-calibrated uncertainty without modifying the base neural network or requiring additional training. Third, we enhance the sensor error model by integrating Gaussian Mixture Models (GMMs), thereby achieving a more precise representation of uncertainty. Fourth, we validate our method in the visual localization problem on a large real-world dataset under various settings and demonstrate that it consistently produces well-calibrated uncertainty estimates and high integrity filters without ad hoc fixes. 


\section{Related Works}
\label{sec:related_works}
\subsection{Uncertainty Modeling.} 
Modern prediction models are known for their high performance in various tasks, but they often lack the ability to tell the uncertainty in their predictions. 
While some models, such as classification neural networks, can produce a confidence score, it is not probabilistic and therefore may not be entirely reliable.
Ensembles~\cite{10.5555/3295222.3295387,ValdenegroToro2019DeepSF,Wen2020BatchEnsembleAA} offer a solution by training multiple networks and combining their predictions to calculate variance and represent uncertainty.
However, ensembles require more costly training steps for training multiple networks, as well as more inference time. 
Bayesian neural networks (BNNs)~\cite{10.1162/neco.1992.4.3.448,10.5555/525544} offer another potential solution by treating neural network weights as random variables instead of deterministic values, with predictions in the form of an expectation over the posterior distribution of the model weights.
Two prominent methods in BNN are Bayes by Backprop~\cite{blundell2015weight} and Monte Carlo (MC) Dropout~\cite{gal2016dropout}. Bayes by Backprop regularises the weights by minimising the expected lower bound on the marginal likelihood.
MC Dropout interprets dropout approximately as integrates over the models’ weights. However, BNN requires specifying a meaningful prior for the parameters which can be challenging.
Additionally, the uncertainty is often poorly calibrated, necessitating post-processing methods~\cite{pmlr-v80-kuleshov18a,guo2017calibration,07998} to map poorly calibrated uncertainty to well-calibrated uncertainty. 
For instance, temperature scaling is a widely used post-processing methods due to its simplicity and effectiveness~\cite{guo2017calibration} . 
 \cite{pmlr-v80-kuleshov18a} extends the technique from just classification tasks to regression tasks.
However, such post-processing methods either require inputs of uncalibrated uncertainty or re-training some layers. 
In contrast, our method differs from these methods in that we do not alter the prediction model's structure, hence preserving its performance. Furthermore, our method can output accurate uncertainty with no additional training and can be applied to any prediction models.

Gaussian mixture models (GMMs) hold potential for application in uncertainty quantification. In the context of non-learning models, GMMs can be integrated into the Gaussian Sum Filter to propagate uncertainty\cite{5152558, 1232327}. This approach is rooted in the notion that any smooth density can be approximated, with some non-zero error, by a GMM featuring a sufficient number of components\cite{Goodfellow-et-al-2016}. For instance, Doe et al.\cite{app13053069}introduced an adaptive GMM that employs virtual sample generation to propagate uncertainty. However, these methods assume that the measurement uncertainty is already known and do not address the issue of modeling the measurement uncertainty, particularly when the measurements are generated from neural networks. In the realm of learning-based models, GMMs can be utilized to quantify uncertainty in neural networks through techniques such as Gaussian Mixture Variational Autoencoder (GMVAE)\cite{dilokthanakul2017deep} or Gaussian Mixture Density Networks (GMDNs)\cite{Bishop1994MixtureDN}. GMVAE leverages a mixture of Gaussian distributions instead of a single Gaussian distribution to model latent variables, allowing for enhanced flexibility and expressiveness. GMDNs, on the other hand, learn the means, covariances, and weights of a Gaussian mixture distribution, with uncertainty expressed via the probabilistic density function of these mixtures. For example, trajectron++\cite{salzmann2021trajectron} incorporated GMMs into their network architecture, enabling multi-modal predictions of vehicle trajectories. However, these approaches often necessitate retraining or modifying the network structure. In contrast, the proposed method in this work stands apart from existing methods by accurately outputting uncertainty without the need for additional training or altering the structure of the prediction model. Notably, this method can be applied to any prediction model while preserving its original structure and performance.

\subsection{Visual Localization}
Visual localization aims to predict the pose of a query image using environmental information such as images and point clouds. 
Two main branches of visual localization are image-based localization and 3D-structure-based localization. 
Image-based localization~\cite{6248018,7410496,Radenovi2016CNNIR} can be understood as an image retrieval problem, \ie retrieving the most similar image from an image database/library with known poses and taking the pose of the retrieved image as the predicted pose. 
Several approaches~\cite{7298790,7780941} have been proposed to extract image features for this purpose 
In contrast, 3D-structure-based localization~\cite{Irschara,8578819,Geppert,Nadeem,Germain,Sattler,8953492} predicts the location by finding the pose that best matches the detected 2D keypoints in the query image with the 3D keypoints in a pre-constructed 3D model.
However, to the best of our knowledge, few works have considered the uncertainty associated with the predicted location. 
While some works~\cite{superglue,superpoint} output confidence scores on detected keypoints and their matching, they do not provide any information about the uncertainty of the predicted location.

\section{Uncertainty Quantification Method}\label{sec:method}

In this section, we discuss our method for uncertainty quantification of prediction models, using visual localization as the application task.
We start by defining a baseline visual localization framework, then present our approach to modeling the errors and calibrating the uncertainties of the predictive network, and finally, we define a full visual localization pipeline with a filter to be used in the validation steps.

 \begin{figure*}[t!]
        \centering 
        \includegraphics[width=0.88\textwidth, trim={0.3cm 3.5cm 0.8cm 0cm},clip]{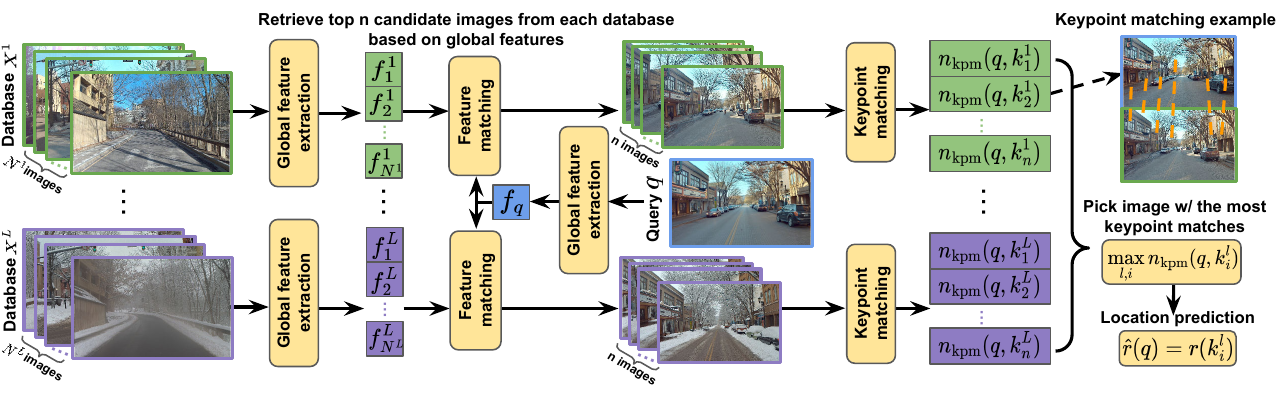}
        \caption{\small Pipeline for location prediction form image-retrieval using multiple traversals.
        \label{fig:vislocpipeline}}
        \vskip-15pt
\end{figure*}

\subsection{Location Prediction from Image Retrieval}\label{sec:visloc}
Let $X=\{k_i\}_{i=1}^N$ be a set of database images with known GPS locations $r(k_i)$.
Given a query image $q$, our goal is to estimate the location where the image was taken.
As images taken from close-by poses should preserve some content similarity, we find the \emph{closest} image $f_\text{closest}(q;X)$ from database $X$ and use its corresponding location as the predicted location $\hat{r}(q)=r(f_\text{closest}(q;X))$. 
We define the \emph{closest} image as the image with the most number of keypoint matches $n_\text{kpm}$ to the query image. However, performing keypoint matching of the query image to all $N$ database images is  computationally expensive. Therefore, more efficient global feature matching (NetVLAD~\cite{7780941}) is performed first, followed by neural keypoint matching (SuperPoint\cite{superpoint} + SuperGlue~\cite{superglue}) on the top $n\!<<\!N$ candidate images. 

\begin{figure}[htbp]
        \vskip-2pt
        \centering
        \includegraphics[width=0.7\linewidth]{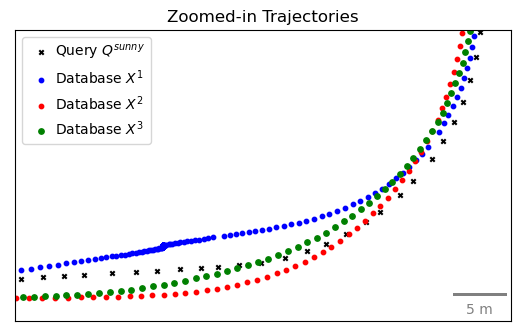}
        \caption{\small GPS locations of several traversals (zoomed in for illustration; full trajectory is not shown). Using multiple traversals increases the chances that a database image is closer to the query image location (\ie, smaller theoretical error).
        \label{fig:motivation1}}
        \vskip-10pt
\end{figure}

A standard location prediction from image retrieval pipeline typically uses a database from just one traversal (passing the route once).
We propose to use \emph{multiple databases} from multiple traversals, motivated by several key observations. First, a query image has a non-zero distance to even its closest image from a database (see \autoref{fig:motivation1}). Using multiple traversals increases the database image options and thus lowers the average error. Second, as the query image for localization can originate from different weather and lighting conditions, it is important to diversify the database images to reduce potential errors (those from the traversal and from keypoint mismatches). Finally and most importantly, data from multiple traversals can be used to provide a localization uncertainty prediction, as will be shown in \ref{sec:uncertainty_pred}.

One naive approach is to simply treat the additional data from multiple traversals $X^1, X^2, \dots, X^L$ as one combined (large) database, and apply the same pipeline. However, this is not effective, as the candidate images retrieved by global feature matching often are biased to come from a single database whose color or even foreground object appearance is most similar to the query image.  
This motivates our new approach that encourages retrieval of candidate images from \textit{each} traversal as shown in \autoref{fig:vislocpipeline}. 
\subsection{Uncertainty Prediction and Quantification}\label{sec:uncertainty_pred}
\subsubsection{Problem Definition} 

Let's consider the probability belief $F(\delta; \psi)$ which represents the probability distribution of the location prediction error variable $\delta \in \mathbb{R}^2$. This error $\delta$ is defined as the difference between the predicted location $\hat{r}=r(f(q;X))$ and the ground truth location $r_{\text{gt}}$, parameterized by the parameters $\psi$.
We formally define the uncertainty quantification problem as predicting the parameters $\psi$.
Our objective is to accurately predict the uncertainty in a manner that is suitable for various levels of confidence. To achieve this, we seek to satisfy the condition: 
\begin{equation}
    p\left(\lVert \hat{r}-r_{\text{gt}} \rVert < \sigma_c \right) = c
\end{equation}
Here, $\sigma_c$ represents the error bound (for a single Gaussian distribution) or contour (for a GMM) corresponding to the confidence level $c$. It means the error between the predicted location $\hat{r}$ and the ground-truth $r_{\text{gt}}$ falls within $\sigma_c$ by $c$ probability.


\subsubsection{Sensor Error Model}
\begin{figure}[htbp]
\vskip-8pt
        \centering
        \includegraphics[width=1\linewidth]{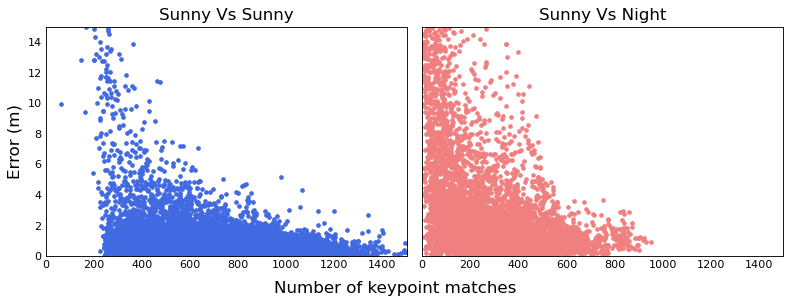}
        \caption{\small Relationship between number of keypoint matches and location error for two different database traversals.}
        \label{fig:motivation3}
        \vskip-8pt
\end{figure}

We propose to create a \textit{sensor error model} to determine the confidence of the prediction (\eg neural network output). A sensor error model maps key attributes of prediction to error bound $\sigma_c$ and confidence $c$ estimates; for example, the error of stereo depth sensor is quadratic to range \cite{gallup2008variable}.
We first analyze the performance of visual localization prediction as a function of the number of keypoint matches $n_\text{kpm}$ by performing cross-validation using different databases. As an example, 
\autoref{fig:motivation3} shows scatter plots of the location error between images from two databases (sunny and night) and their closest images from \emph{another} database (sunny) as a function of the number of keypoint matches.

From this analysis, we learn two things. First, the number of keypoint matches $n_\text{kpm}$ can serve as a good indicator for uncertainty quantification. Second, the relationship between number of keypoint matches and error can be different for different databases (traversals); the scatter plots have different distributions. Thus, we propose to build the sensor error model as a function of number of  keypoint matches, and build one model for each different traversal. We can utilize multiple traversals to learn this mapping as follows.

\subsubsection{Creating Sensor Error Model}
Key to our approach is creating a sensor error model for \textit{each} database/traversal. 
For database $l$, we apply the image retrieval pipeline using traversal $X^l$ as the query and another traversal $X^{m\neq l}$ as the database. For every image $k_i^l \in X^l$, we find the closest image $f(k_i^l; X^m)$ from database $m$ and compute the location error $\lVert r(f(k_i^l; X^m)) - r(k_i^l) \rVert$. Thus, for each image, we can compute the number of keypoint matches (to its closest image) and location error. This process is repeated using all $L-1$ different traversals (other than $X_l$). 
We divide the data (number of keypoints vs error) into  bins according to the number of keypoint matches (\eg, bin 1 contains data points with keypoint matches ranging from 0-200, bin 2 from 200-400, and so on).

For each bin, we provide two distinct approaches to represent the error: one utilizing a single Gaussian distribution, and the other employing Gaussian Mixture Models (GMMs). For single Gaussian distribution,
we empirically determine the error bound $\sigma_c$ for confidence $c$ such that $c$ fraction of data in that bin has smaller error than $\sigma_c$. We repeat for each traversal/database. For GMMs, we fit a Gaussian mixture model to the data within each bin. Subsequent to the derivation of the mixture model, we can computationally ascertain the contour $\sigma_c$ associated with confidence level $c$ for this mixture model. Details can be found in \ref{sec:sensor error model evaluation}.

 \begin{figure*}[t!]
        \centering 
        \includegraphics[width=0.95\textwidth, trim={0.1cm 0cm 0cm 0cm},clip]{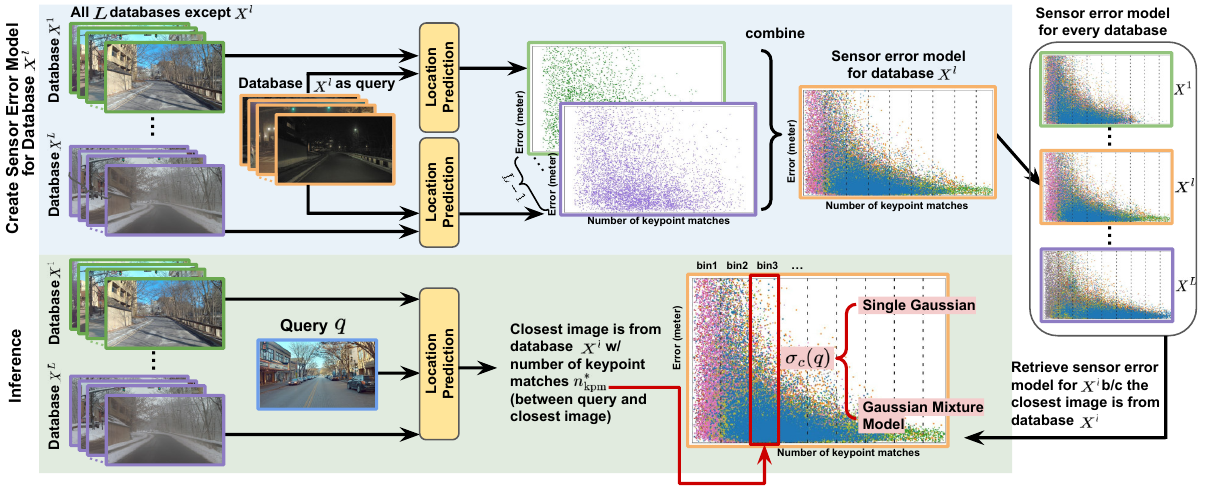}
        \caption{\small Pipeline for uncertainty prediction. Top: creating sensor error model. Bottom: using sensor error model in inference. 
        \label{fig:pipeline}}
        \vskip-15pt
\end{figure*}

\subsubsection{Model Prediction with Uncertainty and Confidence}
The inference process is shown in \autoref{fig:pipeline}(bottom). Given a query image of unknown location, we retrieve the closest image (as detailed in \ref{sec:visloc}); the location of the closest image becomes the predicted location. To find the confidence of the prediction, we use the database of the closest image (say $l$, or $X^l$). The corresponding ($l$) sensor error model is then used; the bin associated with the number of keypoint matches (between query and closest image) gives the corresponding error bound or contour $\sigma_c$ at confidence level $c$.

\subsubsection{Incorporation with Estimation Techniques}
To incorporate with estimation techniques, such as Kalman filters, we form a quantified uncertainty in the form of a 2D estimation error covariance. Specifically, we compute the measurement covariance $R \in \mathbb{R}^{2\times 2}$ from the cross-validation data, per database, per number of keypoint matches range (bin). The covariance matrices are formed and expressed in the ego car (sensor) coordinates. 

To seamlessly integrate the uncertainty represented by Gaussian mixture model into our estimation techniques, we employ the Gaussian Sum Filter (GSF). The GSF enables us to effectively incorporate the GMM-based uncertainty model into our estimation framework.

\subsection{Full Visual Localization Pipeline}\label{sec:method-filter}
\subsubsection{Sigma Point Filter}
We first build a full visual localization pipeline using the location prediction (\ref{sec:visloc}) as the uncertain measurement, the uncertainty prediction represented by single Gaussian distribution (\ref{sec:uncertainty_pred}) as the error covariance, within a formal estimation framework using the Sigma Point (Unscented) filter (SPF)~\cite{wan2000unscented,Brunke2004}. Our goal is to estimate the $p(s_t|m_{1:t})$ of the state vector $s_t$ at time $t$ given observed measurements $m_{1:t}$.
We define the state vector as follows:
\begin{equation}
    s = \begin{bmatrix}x & y & \theta & v & \dot{\theta} \end{bmatrix}^T
\end{equation}
where $x,y,\theta$ are the inertial, planar position and heading angle, and $v, \dot{\theta}$ are the linear and angular velocity of the car. In the prediction step of the SPF, we assume constant linear and angular velocity ($v$ and $\dot{\theta}$) with a small process noise. In the measurement update, given an image input, we process the image through the location and uncertainty prediction pipeline (\ref{sec:visloc} and \ref{sec:uncertainty_pred}) to give the ($x, y$) location measurement and error covariance; the covariance is transformed to the inertial coordinates for the filter. 

\subsubsection{Gaussian Sum Filter}
To seamlessly integrate the sensor error model represented by Gaussian mixture into the visual localization pipeline, we employ the Gaussian Sum Filter (GSF). The GSF represents the probability density function (PDF) of the system state\footnote{states are defined the same with SPF} as a sum of $N_{g_{k}}$ Gaussian components, as shown by the following equation:
\begin{equation}
p(x_{k}|z_{1:k})\approx \sum_{i=1}^{N_{g_{k}}}\omega^{i}_{k}\cdot \mathcal{N}(x^{i}_{k}, P^{i}_{k})
\end{equation}
In the given equation, the variable $k$ denotes the time index and $z$ denotes the measurement. The $\omega$ represents the weight, $\hat{x}$ represents the mean, and $P$ represents the covariance of each Gaussian component. Each Gaussian component corresponds to a potential hypothesis regarding the true state of the system. By performing recursive updates, these components are progressively combined and refined over time, resulting in an increasingly accurate estimation of the state.

The Gaussian Sum Filter (GSF) operates in two main stages at each time step: prediction and update. In the prediction stage, the GSF predicts the new state estimate based on the system dynamics model. Each Gaussian component is propagated according to the model, accounting for any uncertainty or noise in the system. The prediction equation is given by:
\begin{equation}
p(x_{k+1}|z_{1:k})\approx \sum_{i=1}^{N_{g_{k}}\cdot N_{\omega}}\Bar{\omega}^{i}_{k+1}\cdot \mathcal{N}(x^{i}_{k+1}, P^{i}_{k+1})
\end{equation}

In the update stage, the GSF incorporates measurement information to refine the state estimate. In our case, the measurement is represented by a Gaussian Mixture, which is given by:
\begin{equation}
z_{k+1}= \sum_{i=1}^{N_{\upsilon}}\gamma^{i}\cdot \mathcal{N}(\mu^{i}, \Sigma^{i})
\end{equation}
where $\gamma$ is the weight, $\mu$ is the mean and $\Sigma$ is the covariance of each measurement component. The predicted Gaussian components are weighted based on a combination of their previous weights, measurement weights, and their likelihood given the measurement. These weights reflect the confidence or belief in each hypothesis. The updated state estimate equation is:
\begin{equation}
p(x_{k+1}|z_{1:k+1})\approx \sum_{i=1}^{N_{g_{k}}\cdot N_{\omega}\cdot N_{\upsilon}}\Tilde{\omega}^{i}_{k+1}\cdot \mathcal{N}(x^{i}_{k+1}, P^{i}_{k+1})
\end{equation}
\begin{equation}
\Tilde{\omega}^{i}_{k+1}=\Bar{\omega}^{j}_{k+1}\gamma^{l}p(\hat{z}^{l}_{k+1}|x_{k+1}^{j}) 
\end{equation}
where $\Bar{\omega}^{j}_{k+1}$ stands for each component weight in $p(x_{k+1}|z_{1:k})$ and $\gamma^{l}$ stands for each component weight in $z_{k+1}$.  

After the update stage, the GSF performs a pruning and merging step to maintain a tractable number of Gaussian components while still representing the underlying PDF accurately. Following Runnalls’s\cite{4383588} condensation algorithm, given two weighted
Gaussian densities $\omega^{i}\cdot \mathcal{N}(x^{i}, P^{i})$ and $\omega^{j}\cdot \mathcal{N}(x^{j}, P^{j})$ from a Gaussian mixture, the merged weight $\omega^{ij}$, mean $x^{ij}$ and covariance $P^{ij}$ are:
\begin{align}
\omega^{ij}&=\omega^{i}+\omega^{j} \\
x^{ij} &= \frac{\omega^{i}}{\omega^{ij}}x^{i}+\frac{\omega^{j}}{\omega^{ij}}x^{j}\\
P^{ij}&=\frac{\omega^{i}}{\omega^{ij}}P^{i}+\frac{\omega^{j}}{\omega^{ij}}P^{j}+\frac{\omega^{i}\omega^{j}}{(\omega^{ij})^2}(x^{i}-x^{j})(x^{i}-x^{j})^{T}
\end{align}
The condensation approach in the Gaussian Sum Filter (GSF) iteratively merges components based on a bounded approximation of the Kullback-Leibler divergence between the original and reduced mixture. This divergence $B^{ij}$ is bounded by the expression:

\begin{equation}
\begin{aligned}
B^{ij} = 0.5 \cdot \Big[ &(w^{i} + w^{j})\log\det(P^{ij}) \\
&- w^{i}\log\det(P^{i}) - w^{j}\log\det(P^{j}) \Big]
\end{aligned}
\end{equation}

In the GSF, pruning is applied to remove Gaussian components with low weights, indicating their reduced relevance or unlikelihood given the measurements. On the other hand, merging combines Gaussian components that are close together, reducing the number of components without significant loss of information. Through the iterative process of prediction, update, pruning, and merging, the GSF dynamically adapts to the evolving characteristics of the system and measurements. This adaptation allows it to provide an estimate of the system state that incorporates both the dynamics model and the information from the measurements. 

During the experiment, we opt for a configuration where the state is represented using two mixtures, while the measurement uncertainty is captured by three mixtures. Our findings indicate that employing a greater number of mixtures does not yield a substantial improvement in performance.



\section{Experiments}
\label{sec:experiments}
\subsection{Dataset}
We use the Ithaca365 dataset~\cite{Diaz-Ruiz_2022_CVPR}, 
containing data collected over multiple traversals along a 15km route under various conditions: snowy, rainy, sunny, and nighttime. We utilize two types of sensor data, images, and GPS locations for our experiments. 
 For our database, we randomly select nine traversals, with three traversals each from the sunny  ($X^1$, $X^2$, $X^3$), nighttime ($X^4$, $X^5$, $X^6$), and snowy ($X^7$, $X^8$, $X^9$). 
 We use three additional traversals ($Q^\text{sunny}$, $Q^\text{snow}$ and $Q^\text{night}$), one from each condition, as queries for testing and evaluation.
 To avoid double counting and ensure a uniform spatial distribution across the scenes in evaluation, we sample query images at an interval of $\approx$1m, except for highways where the spacing is larger. 
This results in an average of  $\approx$10,000 images for each query traversal, $Q^{(\cdot)}$. 


\subsection{Sensor Error Model Evaluation}\label{sec:sensor error model evaluation} 
First, we evaluate the correctness of our uncertainty prediction on \emph{location prediction using image retrieval}. 
Following~\cite{pmlr-v80-kuleshov18a,guo2017calibration}, we use \textit{reliability diagram} and \textit{Expected Calibration Error (ECE)} to compare the expected confidence level with the observed confidence level.

\begin{figure}[htb]
    \begin{minipage}[t]{1\linewidth}
        \centering
        \includegraphics[width=1\linewidth]{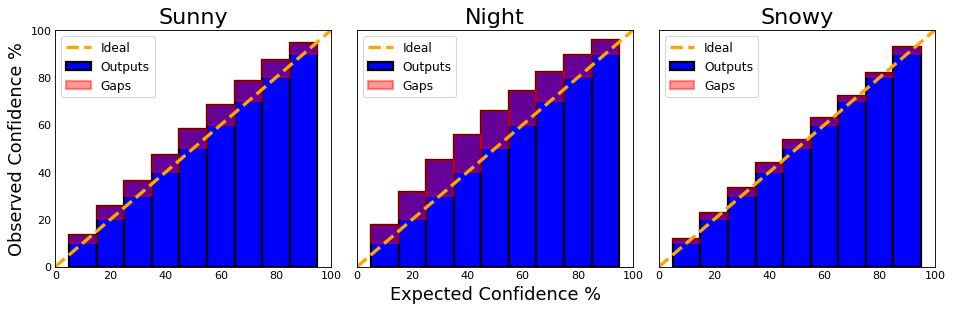}
    \end{minipage}  
    \caption{\small Reliability diagrams of Gaussian sensor error model for $Q^\text{sunny}$, $Q^\text{night}$, and $Q^\text{snowy}$.}
    \label{fig:eval_measurement_Gau}
\end{figure}

First let us begin by introducing the evaluation procedure for a \emph{single Gaussian} sensor error model. For a \emph{single Gaussian} sensor error model, given one expected confidence level $c$, the observed confidence can be obtained by computing the empirical frequency $\hat{p}_{c}$ that the location error $\lVert\hat{x}(q)-x_{\text{gt}}(q) \rVert$ is below the predicted uncertainty $\sigma_{c}(q)$. We can extend the same concept for evaluating \emph{GMM} sensor error model. For a 2-dimensional ($xy$) \emph{GMM} sensor error model, given an expected confidence level $c$, we can calculate the observed confidence as follows. First, we find the contour line(s)\footnote{This contour line is a line with the same probability density function} $\sigma_{c}$ such that the total probability integration inside the contour area(s)\footnote{Note that it is possible to have multiple closed contour lines, hence multiple separate contour areas.} $\mathcal{A}_c$ sums to $c$. Then, the observed confidence  for GMM can be obtained by computing the empirical frequency $\hat{p}_c$ where the deviation from the groundtruth $\hat{x}(q)-x_{\text{gt}}(q) \in \mathbb{R}^2$ falls inside the area $\mathcal{A}_c$. With a slight abuse of notation, the two evaluations can be summarized by the equation below: 
\begin{equation}\label{eqn:observed freq GMM}
\hat{p}_{c}=\frac{\left|\{q \in Q\ \text{ s.t. }  \space \lVert\hat{x}(q)-x_{\text{gt}}(q) \rVert\leq\sigma_{c}(q)\}\right|}{|Q|}.
\end{equation}
The reliability diagram (as shown in \autoref{fig:eval_measurement_Gau}) then plots the expected confidence ($c$) versus the observed confidence ($\hat{p}_c$).
If the uncertainty quantification is accurate, the reliability diagram should plot the identity function (a straight line with a gradient of one).

\begin{figure}[htb]
    \begin{minipage}[t]{1\linewidth}
        \centering
        \includegraphics[width=1\linewidth]{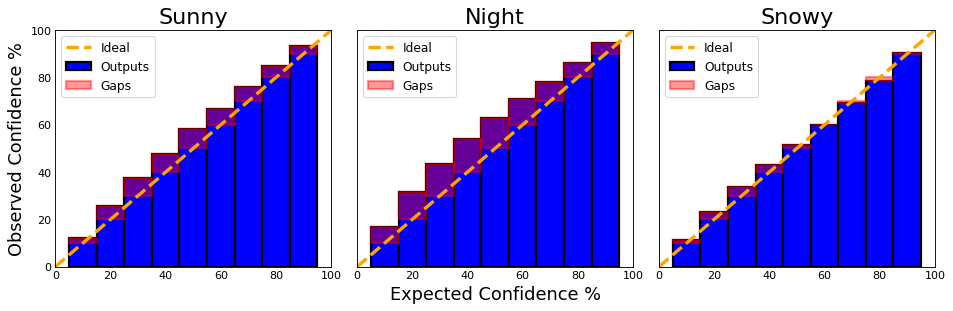}
    \end{minipage}  
    \caption{\small Reliability diagrams of GMM sensor error model for $Q^\text{sunny}$, $Q^\text{night}$, and $Q^\text{snowy}$.}
    \label{fig:eval_measurement_GMM}
\end{figure}

While the reliability diagram serves as a useful visual tool, having a single numerical summary of the statistic can offer benefits and simplify the comparison process, especially when dealing with a multitude of evaluations. 
Thus, we can summarize the reliability diagram using ECE (Expected Calibration Error) defined as the absolute probability error ($|\hat{p}_c - c|$) averaged across different confidence levels ($c$). In other words, we take the absolute gaps shown in the reliabilty diagram (\autoref{fig:eval_measurement_Gau} and \autoref{fig:eval_measurement_GMM}) and average them across all the bins.
ECE serves as a scalar summary statistic derived from the reliability diagram to measure how well the predicted probabilities match the true (observed) probabilities. 

The reliability diagram in \autoref{fig:eval_measurement_Gau} and \autoref{fig:eval_measurement_GMM} shows that both of our method produce accurate probabilistic confidence, as evidenced by the small gaps between observed and expected confidence at all levels and across all three conditions. 



\begin{figure}[htb]
  \centering
  \begin{subfigure}{0.2\textwidth}
    \includegraphics[width=\linewidth]{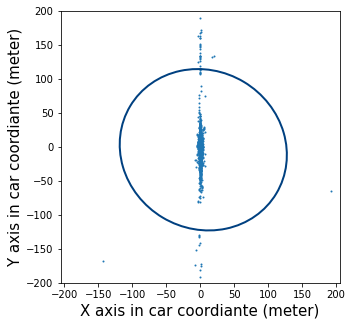}
    \caption{Single Gaussian}
    \label{fig:gau_subfiga}
  \end{subfigure}
  \hfill
  \begin{subfigure}{0.2\textwidth}
    \includegraphics[width=\linewidth]{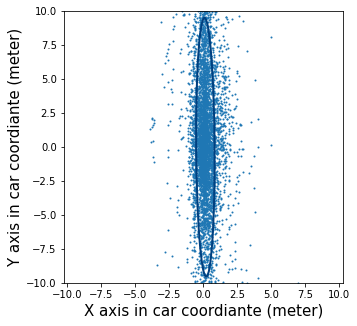}
    \caption{Gaussian Mixture 1}
    \label{fig:gau_subfigb}
  \end{subfigure}
  \hfill
  \begin{subfigure}{0.2\textwidth}
    \includegraphics[width=\linewidth]{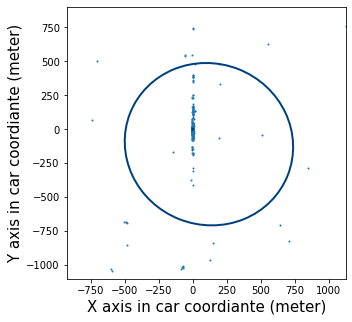}
     \caption{Gaussian Mixture 2}
    \label{fig:gau_subfigc}
  \end{subfigure}
  \caption{\small Comparison between an oversized single Gaussian and Gaussian Mixtures on error data. The left figure shows data fitted by a single Gaussian, which is oversized. The right two figures display Gaussian mixtures of two mixands fitting on the same data.}
  \label{fig:oversized}
\end{figure}

We present the ECE evaluation results for both the \emph{single Gaussian} and \emph{GMM} error models in \autoref{table:eval_ECE}. Our ECE analysis reveals that the GMM sensor error model, utilization of Gaussian Mixture Model (GMM) for fitting error data, consistently outperforms Gaussian sensor error model, the use of a single Gaussian distribution for fitting error data. The advantage of employing GMMs lies in their ability to capture the underlying complexity of the error distribution more effectively. The limitation of relying solely on a single Gaussian distribution becomes evident when attempting to model intricate error distributions. For instance, consider the scenario depicted in \autoref{fig:gau_subfiga} (several data points fall outside the range of the axis), where attempting to describe a multifaceted distribution with a single Gaussian model leads to an oversized and inadequately fitted curve. In contrast, the flexibility inherent in Gaussian mixtures enables them to provide a more accurate representation of such distributions, as demonstrated in \autoref{fig:gau_subfigb} and \autoref{fig:gau_subfigc}. These findings underscore the significance of considering Gaussian Mixture Models when addressing the challenge of modeling error distributions such as under night and snowy conditions. By accommodating the inherent complexities within the data, GMMs contribute to enhanced accuracy and a more faithful representation of the underlying uncertainty. 

\subsection{Visual Localization Evaluation: Filter + Prediction/Error Model} 
\label{sec:SPFfilter}
Next, we evaluate the \emph{full visual localization pipeline}, which uses the previous location predictions as measurements in the Sigma Point Filter (SPF) and Gaussian Sum Filter (GSF) (\autoref{sec:method-filter}).
We evaluate both the localization error and the uncertainty of the estimates. The localization error $d_\text{err}$ is defined as the average  distance error between estimated  and ground-truth locations. 
To assess the accuracy of the uncertainty or distribution belief associated with the state estimate of the filter, we report the \emph{covariance credibility}.
 Specifically, the covariance credibility quantifies the observed confidence for various expected confidence levels (\ie, $c=68\%$, $95\%$ and $99.7\%$). 
Ideally, the observed confidence values should match the expected values.
By comparing the observed confidence with the expected confidence levels, we gain insights into the precision of the uncertainty estimation. This concept parallels the uncertainty evaluation described in \autoref{sec:sensor error model evaluation}, but with a focus on evaluating the predicted state uncertainty estimate within the filter instead of the measurement model.

In the case of the \emph{SPF}, which employs \emph{single Gaussians} to represent both the measurement model and the state estimate, the observed confidence can be expressed in a simple form: the observed frequency at which the filter 2D localization error falls within an $n$-sigma covariance ellipse. Here, we consider 1-, 2-, and 3-sigma levels, corresponding to the expected probabilities of $68\%$, $95\%$ and $99.7\%$ in a 2D Gaussian distribution.

Extending this concept to the \emph{GSF}, where the measurement model and state estimate are characterized using \emph{Gaussian Mixture Model (GMM)}, follows a similar procedure. We determine the observed confidence for $68\%$, $95\%$ and $99.7\%$ for the filter state belief. This process mirrors \autoref{eqn:observed freq GMM}, but now we evaluate the \emph{state estimate belief} derived from the filter, rather than the measurement model.
Given the belief distribution of the state estimate, we identify the contour line(s) $\mathcal{L}_c$ such that the total probability within the contour area(s) $\mathcal{A}_c$ equals $c$. 
Subsequently, we determine the observed frequency at which the 2D localization error of the state estimate falls within these contours, in accordance with \autoref{eqn:observed freq GMM}. This observed confidence, termed as covariance-credibility with a slight abuse of term, is then reported as a measure of the accuracy of the state uncertainty estimate.

\begin{figure}[htbp]
\vskip-5pt
        \centering
        \includegraphics[width=0.78\linewidth]{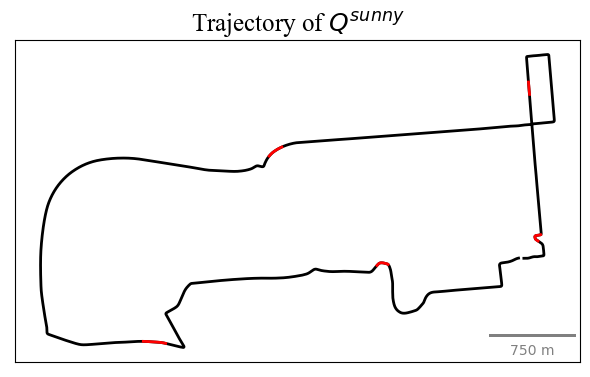}
        \caption{\small Data collection path (black), with corrupted images (red). 
        \label{fig:studycase-map1}}
        \vskip2pt
        \centering
        \includegraphics[width=0.9\linewidth]{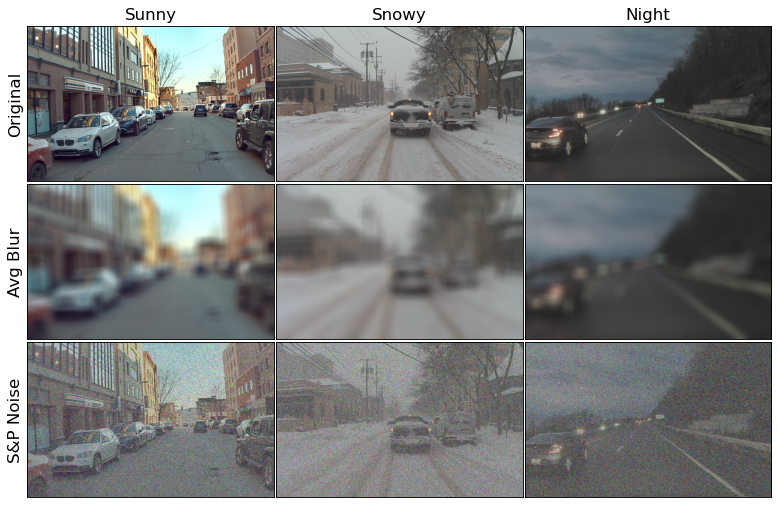}
        \caption{\small Examples of corrupted images. Top: original images. Mid: blurred images (average blur with kernel size 80, Bottom: images corrupted with salt and pepper noise (noise amount is 0.5)).
        \label{fig:studycase-imgmanipulation}}
        \vskip-5pt
\end{figure}

We present three sets of experiments in \autoref{table:eval_UKF_car}. 
The first set of experiments (rows 1-9) uses the original image inputs. The second and the third sets simulate high sensor error/failure by corrupting \emph{several} images along the \emph{red paths} of \autoref{fig:studycase-map1}.
Specifically, the second set (rows 10-15) applies average blurring, and the third set (rows 16-21) applies salt and pepper noise, as shown in \autoref{fig:studycase-imgmanipulation}. 

We benchmark our method against the \emph{constant covariance baselines}, widely employed in Kalman filters.
The baseline+SPF employs a \emph{constant} GMM (hence, constant weights and mixand components) to represent the measurement error model, where the constant covariance value is obtained from individual tuning of the validation data, for each distinct weather condition.
On the other hand, the baseline+GSF employs a \emph{constant}
Our method and the constant covariance baselines receive the \emph{same measurement vectors} but use \emph{different measurement uncertainty}. 
Additionally, in the first experiment set, we provide a comparison to the Monte Carlo (MC) Dropout method. Specifically, we apply a dropout layer after the final keypoint feature projection layer with a 0.3 dropout probability and repeat the dropout process multiple times until the SPF localization results stabilize. We report the converged results.

Analysis of \autoref{table:eval_UKF_car} reveals several observations. Firstly, our method (Gaussian+SPF and GMM+GSF) outperforms the MC Dropout and constant covariance baselines in terms of localization accuracy ($d_\text{err}$) in nearly all cases, suggesting that a good uncertainty model can improve localization accuracy, even with similar measurement quality. Our method also produces more accurate uncertainty estimates (indicated by covariance-credibility) than the baselines in nearly all cases. This is crucial for making informed decisions in the future. Second, the uncertainty represented by GMM performs better than uncertainty repersented by single Gaussian, especially in challenging conditions, such as night, where the environment exhibits greater complexity. By effectively capturing the multiple modes inherent in the error data, GMM+GSF significantly enhances localization accuracy under such circumstances. 
 \setlength{\tabcolsep}{2.5pt}
\begin{table}[!bt]
\centering
\caption{Evaluation results of expected calibration error. Lower is better.}
\label{table:eval_ECE}
\begin{tabular}{lccc}
\toprule

Sensor error model & sunny & night & snow \\
\midrule 
Gaussian & 6.70\% & 11.98\% & 2.80\%   \\
GMM & 5.77\% & 9.70\% & 1.72\%   \\

\bottomrule
\end{tabular}
\end{table}

 \setlength{\tabcolsep}{2.5pt}
\begin{table*}[!bt]
\centering
\begin{center}
\caption{Evaluation result of Kalman Filter localization for \cone{sunny}, \ctwo{night}, and \cthree{snowy} conditions.}
\label{table:eval_UKF_car}
\begin{tabular}{cl@{\hskip 0.8in}cc@{\hskip 0.4in}cc@{\hskip 0.4in}cc}
\toprule
Row & Method &  \cone{$d_\text{err}$(m) $\downarrow$}& \cone{Cov-credibility(\%)}  & \ctwo{$d_\text{err}$(m) $\downarrow$} & \ctwo{Cov-credibility(\%)}  & \cthree{$d_\text{err}$(m) $\downarrow$} & \cthree{Cov-credibility(\%)} \\ \midrule 
\multicolumn{8}{c}{Normal measurement --- Average measurement error: \cone{0.83m} / \ctwo{11.67m} / \cthree{1.76m} for MC dropout -- \cone{0.87m} / \ctwo{8.68m} / \cthree{1.42m} for baseline and ours.}
\\
\midrule
1 & MC d.o &  \cone{0.79} & \cone{32.8 / 67.4 / 84.7}  & \ctwo{7.44} & \ctwo{27.6 / 56.2 / 71.0}  & \cthree{1.50} & \cthree{20.4 / 50.4 / 69.8} \\
2 & Baseline+SPF &  \cone{0.76} & \cone{46.1 / 81.2 / 92.3}  & \ctwo{8.42} & \ctwo{38.5 / 67.0 / 78.9}  & \cthree{1.30} & \cthree{30.5 / 61.8 / 79.6} \\
3 & Gaussian+SPF     & \cone{0.57} & \cone{62.2 / 91.4 / 97.4}  & \ctwo{3.07} & \ctwo{75.9 / \textbf{94.6} / \textbf{98.6}} &  \cthree{\textbf{0.81}} & \cthree{55.4 / 86.4 / 95.9} \\ 
4 & Baseline+GSF      & \cone{1.06} & \cone{97.3 / 99.0 / \textbf{99.2}}  & \ctwo{1.13} & \ctwo{\textbf{66.8} / 88.8 / 96.2} &   \cthree{1.21} & \cthree{52.3 / 82.7 / 94.2} \\
5 & GMM+GSF      & \cone{\textbf{0.57}} & \cone{\textbf{70.7} / \textbf{93.4} / 98.5}  & \ctwo{\textbf{1.09}} & \ctwo{73.9 / 93.9 / 98.5}  & \cthree{0.84} & \cthree{\textbf{61.8} / \textbf{88.1} / \textbf{96.8}} \\

\midrule
\multicolumn{8}{c}{Study case where \cone{5.0\%} / \ctwo{4.7\%} / \cthree{4.6\%} of data are  corrupted with average blurring --- Average measurement error: \cone{108.21m} / \ctwo{121.78m} / \cthree{108.66m}}\\
\midrule
10 & Baseline+SPF  & \cone{107.16} & \cone{41.9 / 74.6 / 85.4}   & \ctwo{121.37} & \ctwo{36.3 / 63.5 / 74.4}  & \cthree{108.08} & \cthree{29.2 / 58.9 / 75.3} \\
11 & Gaussian+SPF      & \cone{3.37} & \cone{61.0 / 90.3 /  96.9} & \ctwo{7.70} & \ctwo{\textbf{72.9} / 92.7 / 97.2}  & \cthree{3.35} & \cthree{56.4 / 87.0 / 95.7} \\ 
12 & Baseline+GSF      & \cone{120.70} & \cone{86.1 / 88.2 /  89.3}  & \ctwo{107.19} & \ctwo{44.6 / 60.4 / 68.3} & \cthree{109.64} & \cthree{49.5 / 77.3 / 87.5} \\
13 & GMM+GSF      & \cone{\bf{2.83}} & \cone{\bf{70.0} / \bf{92.2} / \bf{98.2}}  & \ctwo{\textbf{4.13}} & \ctwo{75.4 / \bf{93.0} / \bf{97.7}}  & \cthree{\textbf{3.29}} & \cthree{\textbf{63.2} / \textbf{88.3} / \textbf{96.5}} \\ 

\midrule
\multicolumn{8}{c}{Study case where \cone{5.0\%} / \ctwo{4.7\%} / \cthree{4.6\%} of data are corrupted with salt and pepper noise --- Average measurement error:  \cone{66.09m} / \ctwo{103.63m} / \cthree{85.26m}}\\
\midrule
16 & Baseline+SPF  & \cone{65.36} & \cone{42.0 / 74.9 / 85.6} & \ctwo{102.88} & \ctwo{36.8 / 63.9 / 75.0}  & \cthree{84.73} & \cthree{29.1 / 58.7 / 75.1} \\
17 & Gaussian+SPF     & \cone{1.97} & \cone{61.9 / 91.1 / 97.4}  & \ctwo{7.08} & \ctwo{\textbf{73.1} / 92.2 / 97.1} & \cthree{\textbf{2.46}} & \cthree{57.0 / 87.4 / 96.1} \\ 
18 & Baseline+GSF      & \cone{74.36} & \cone{88.1 / 89.7 / 90.1}  & \ctwo{87.01} & \ctwo{42.7 / 62.1 / 72.6}  & \cthree{84.55} & \cthree{49.6 / 77.4 / 87.7} \\
19 & GMM+GSF      & \cone{\bf{1.44}} & \cone{\bf{69.9} / \bf{92.3} / \bf{98.0}}  & \ctwo{\textbf{3.59}} & \ctwo{75.5 / \bf{93.1} / \bf{98.1}} & \cthree{2.55} & \cthree{\textbf{63.3} / \textbf{88.7} / \textbf{97.0}} \\ 


\bottomrule
\end{tabular}
\end{center}
\vskip-10pt
\end{table*}
 \setlength{\tabcolsep}{1.4pt}

\subsection{Generalization Test}
\begin{figure}[htbp]
\vskip-5pt
        \centering
        \includegraphics[width=0.78\linewidth]{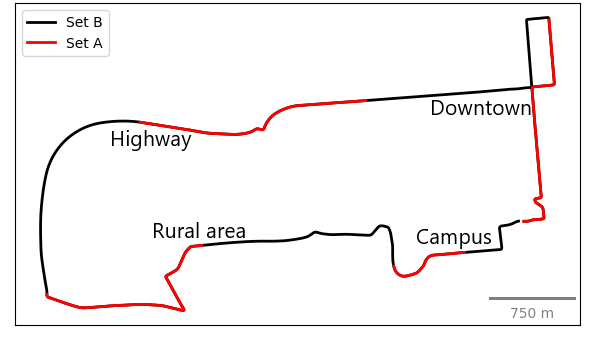}
        \caption{\small Data split.  
        \label{fig:split_map}}
        \vskip-5pt
\end{figure}

In practice, one may not want to collect multiple traversals of data from every city to construct the sensor error model. 
One possible solution is to create a single sensor error model from one location and apply it to other locations.
In the following, we evaluate the feasibility of using a sensor error model constructed from one location and applying it to a different location. 
We partition our available data into two sets: Set A and Set B (illustrated in red and black in \autoref{fig:split_map}).
We utilize multiple traversals of data from Set B to build the sensor error model. 
Subsequently, we deploy the sensor error model from B on a different location (Set A), which were not used for constructing. It's important to mention that the data is further divided into four smaller chunks (as seen in \autoref{fig:split_map}) each for Set A and Set B. This division is deliberate to ensure that the evaluation encompasses diverse scenes (\ie, downtown, highway, rural areas, and campus area) in order to maintain representativeness. For this experiment, we use the single Gaussian sensor error model to represent the uncertainty. We use the same location data (Set A) to create a sensor error model as a baseline. 


\subsubsection{Sensor Error Model}
Similar as our previous evaluation (\autoref{sec:sensor error model evaluation}), we first evaluate the correctness of our uncertainty prediction on \emph{location prediction using image retrieval}. 
We present the ECE results in \autoref{table:eval_unseen_ECE}.
We observe that utilizing either data B or data A for sensor error model construction exhibit similar levels of performance. This finding strongly suggests that our model exhibits substantial generalization of the sensor error model when applied to unseen locations.

\subsubsection{Visual Localization: Filter + Prediction/Error Model}

 \setlength{\tabcolsep}{2.5pt}
\begin{table}[!bt]
\centering
\caption{Evaluation results of expected calibration error for unseen locations (set A). Lower is better.}
\label{table:eval_unseen_ECE}
\begin{tabular}{clcccc}
\toprule

Condition & Sensor error model & Downtown & Highway& Rural area & Campus \\
\midrule
\multirow{2}{*}{Sunny} & Using Set A & 10.33\% & 10.86\% & 0.80\% &11.21\%    \\
& Using Set B & 5.64\% &  9.58\% &  2.73\% &  9.30\%   \\
\midrule
\multirow{2}{*}{Night} & Using Set A & 4.23\% & 15.22\% & 8.13\% & 10.80\%     \\
& Using Set B & 1.26\% & 15.97\% &7.36\% & 5.34\%    \\
\midrule
\multirow{2}{*}{Snowy} & Using Set A & 1.71\% & 4.28\% & 2.94\% & 8.80\%    \\
 & Using Set B & 5.67\% &  5.06\% & 1.43\% & 8.38\%     \\

\bottomrule
\end{tabular}
\end{table}

 \setlength{\tabcolsep}{2.5pt}
\begin{table}[!bt]
\centering
\caption{Evaluation results of Sigma Point Filter localization for unseen location (set A).}
\label{table:eval_UKF_unseen}
\begin{tabular}{cclcc}
\toprule
Condition & Area & Method & $d_\text{err}$(m) $\downarrow$ & Cov-credibility(\%) \\ 
\midrule

\multirow{8}{*}{Sunny} & \multirow{2}{*}{Downtown} & Using Set A  & 0.46 & 
60.9 / 92.5 / 98.4
  \\
& & Using Set B & 0.46 & 63.9 / 93.4 / 98.5
  \\
\cmidrule{2-5}

 & \multirow{2}{*}{Highway} & Using Set A  & 0.57 &
73.8 / 95.6 / 99.1
  \\
& & Using Set B & 0.56 & 73.4 / 95.1 / 99.0
  \\
\cmidrule{2-5}

 & \multirow{2}{*}{Rural area} & Using Set A  & 0.73 &
49.9 / 84.6 / 95.5
  \\
& & Using Set B & 0.72 & 52.1 / 85.2 / 96.2
  \\
\cmidrule{2-5}

 & \multirow{2}{*}{Campus} & Using Set A  & 0.55 & 
67.0 / 95.7 / 99.4
  \\
& & Using Set B & 0.54 & 69.0 / 96.0 / 99.7
  \\
\midrule

\multirow{8}{*}{Night} & \multirow{2}{*}{Downtown} & Using Set A  & 0.61 &
51.1 / 85.1 / 97.2
  \\
& & Using Set B & 0.61 & 53.0 / 86.2 / 97.7

  \\
\cmidrule{2-5}

& \multirow{2}{*}{Highway} & Using Set A  & 2.58 & 
89.1 / 99.4 / 99.9
  \\
& & Using Set B & 1.36 & 90.6 / 99.4 / 99.9
  \\
\cmidrule{2-5}

& \multirow{2}{*}{Rural area} & Using Set A  & 5.54 & 
81.0 / 96.4 / 99.2
  \\
&  & Using Set B & 5.56 & 71.3 / 92.9 / 96.8

  \\
\cmidrule{2-5}

& \multirow{2}{*}{Campus} & Using Set A  & 0.93 &
60.2 / 98.2 / 100.0

  \\
& & Using Set B & 0.84 & 60.8 / 99.1 / 99.8 \\

\midrule

\multirow{8}{*}{Snowy} & \multirow{2}{*}{Downtown} & Using Set A  & 0.78 &
42.4 / 78.5 / 93.1
  \\
& & Using Set B & 0.84 & 60.8 / 99.1 / 99.8
  \\
\cmidrule{2-5}

 & \multirow{2}{*}{Highway} & Using Set A  & 1.18 &
44.3 / 78.3 / 93.4
  \\
& & Using Set B & 1.19 & 44.8 \ 77.7 \ 92.6

  \\
\cmidrule{2-5}

 & \multirow{2}{*}{Rural area} & Using Set A & 0.88&
52.6 / 85.2 / 95.2

  \\
& & Using Set B & 0.87 & 54.8 / 85.0 / 95.5

  \\
\cmidrule{2-5}

 & \multirow{2}{*}{Campus} & Using set B  & 0.66 & 
60.0 / 91.7 / 98.2

  \\
&  & Using Set B & 0.66 & 60.3 / 92.2 / 97.7

  \\
\bottomrule
\end{tabular}
\end{table}

Next, we evaluate the \emph{full visual localization pipeline} (similar as \autoref{sec:SPFfilter}). As we employ a single Gaussian model for uncertainty representation in this experiment, thus we employ the SPF for the filter. We evaluate both the localization error ($d_{err}$) and the uncertainty of the estimates (cov-credibility). We report the results in \autoref{table:eval_UKF_unseen}.
Once again, we observe that employing either sensor error model from different location (data B) or same location (data A)  results in similar levels of performance in both the localization error and uncertainty evaluation. 
This observation further illustrates the feasibility of employing a sensor error model from one location and applying it in a different location. 
This characteristic enhances the scalability and ease of adoption of our approach, as a single sensor error model can be effectively utilized across various new locations.


\subsection{Latency and Data Size}
On a 1080Ti GPU, extracting global features of an image using NetVLAD takes about 8ms, while performing keypoint matching for a single pair of images using SuperPoint and SuperGlue takes approximately 112ms. Although keypoint matching is done between a query image and ten candidate images, the GPU can simultaneously process them in a batch without affecting the speed. The database comprises 127,225 images with a total size of 417.7 GB. Instead of storing the original images, we only need to store the extracted global features (2.24GB) and the keypoint features (161.9GB).

\section{Conclusion}

We present a general and formal probabilistic approach for modeling prediction (\eg, neural network) uncertainties, which we validate in the context of visual localization problem. 
Our approach involves creating a sensor error model that maps the output of the internal prediction model (number of keypoint matches) to probabilistic uncertainty for each database. We delve into two distinct techniques for expressing this uncertainty: the first involves a single Gaussian distribution, while the second employs Gaussian Mixture Models. 
During inference, we use the sensor error model to map the number of keypoint matches to confidence probability.  
We evaluate our approach using a large-scale real-world  self-driving dataset with varying weather, lighting, and sensor corruption conditions, demonstrating accurate uncertainty predictions across all conditions. Particularly noteworthy is the superior performance of Gaussian Mixture Models in representing uncertainty, particularly in challenging and adverse environments. Our approach results in more robust and  better-performing perception pipelines.



\section*{Acknowledgement}
This research is supported by grants from the National Science Foundation NSF (CNS-2211599 and IIS-2107077) and the ONR MURI (N00014-17-1-2699).










{\small
\bibliographystyle{unsrt}
\bibliography{ICRAuncertainty}
}

\end{document}